\documentclass[10pt,twocolumn,letterpaper]{article}

\usepackage{iccv}
\usepackage{times}
\usepackage{epsfig}
\usepackage{subfigure}
\usepackage{booktabs}
\usepackage{multirow}
\usepackage{url}
\usepackage[T1]{fontenc}
\usepackage[utf8]{inputenc}
\usepackage{graphicx}
\usepackage{amsmath,amssymb} 
\usepackage{color}
\usepackage{algorithm}
\usepackage[noend]{algpseudocode}
\usepackage{arydshln}
\usepackage{amsthm}
\usepackage{scrextend}

\def\mb{\mathbf}

\def\etal{\emph{et al.}}
\def\eg{\emph{e.g., }}
\def\ie{\emph{i.e., }}
\def\wrt{\emph{w.r.t. }}
\def\etc{\emph{etc.}}

\usepackage[pagebackref=true,breaklinks=true,letterpaper=true,colorlinks,bookmarks=false]{hyperref}

\iccvfinalcopy 

\def\iccvPaperID{1859} 

\ificcvfinal\fi
\begin{document}

\title{Layout-induced Video Representation for \\Recognizing Agent-in-Place Actions}



\author{Ruichi Yu$^{1}$ ~~~Hongcheng Wang$^{2}$ ~~~Ang  Li$^3$\thanks{This work was done while the author was at the University of Maryland.}
~~~Jingxiao Zheng$^1$
~~~ Vlad I. Morariu$^{4*}$
~~~ Larry S. Davis$^1$\\
$^1$University of Maryland, College Park   ~~~~~~~~~~$^2$Comcast Applied AI Research\\
~~~~~~~~~~ $^3$DeepMind  ~~~~~~~~~~$^4$Adobe Research \\
{\tt\small {\{yrcbsg, jxzheng, lsd\}@umiacs.umd.edu},  hongcheng\_wang@comcast.com
}
\\
{\tt\small {anglili@google.com, morariu@adobe.com}
}
}

\maketitle

\begin{abstract}
We address scene layout modeling for recognizing agent-in-place actions, which are actions associated with \textit{agents} who perform them and the \textit{places} where they occur, in the context of outdoor home surveillance. 
We introduce a novel representation to model the geometry and topology of scene layouts so that a network can generalize from the layouts observed in the training scenes to unseen scenes in the test set. 
This Layout-Induced Video Representation (LIVR) abstracts away low-level appearance variance and encodes geometric and topological relationships of places to explicitly model scene layout.
LIVR partitions the semantic features of a scene into different places to force the network to learn generic place-based feature descriptions which are independent of specific scene layouts;
then, LIVR dynamically aggregates features based on connectivities of places in each specific scene to model its layout.
We introduce a new Agent-in-Place Action (APA) dataset\footnote{The dataset is pending legal review and will be released upon the acceptance of this paper.} to show that our method allows neural network models to generalize significantly better to unseen scenes. 
\end{abstract}

\section{Introduction}

\begin{figure}[t]
\centering
  \includegraphics[width=\linewidth]{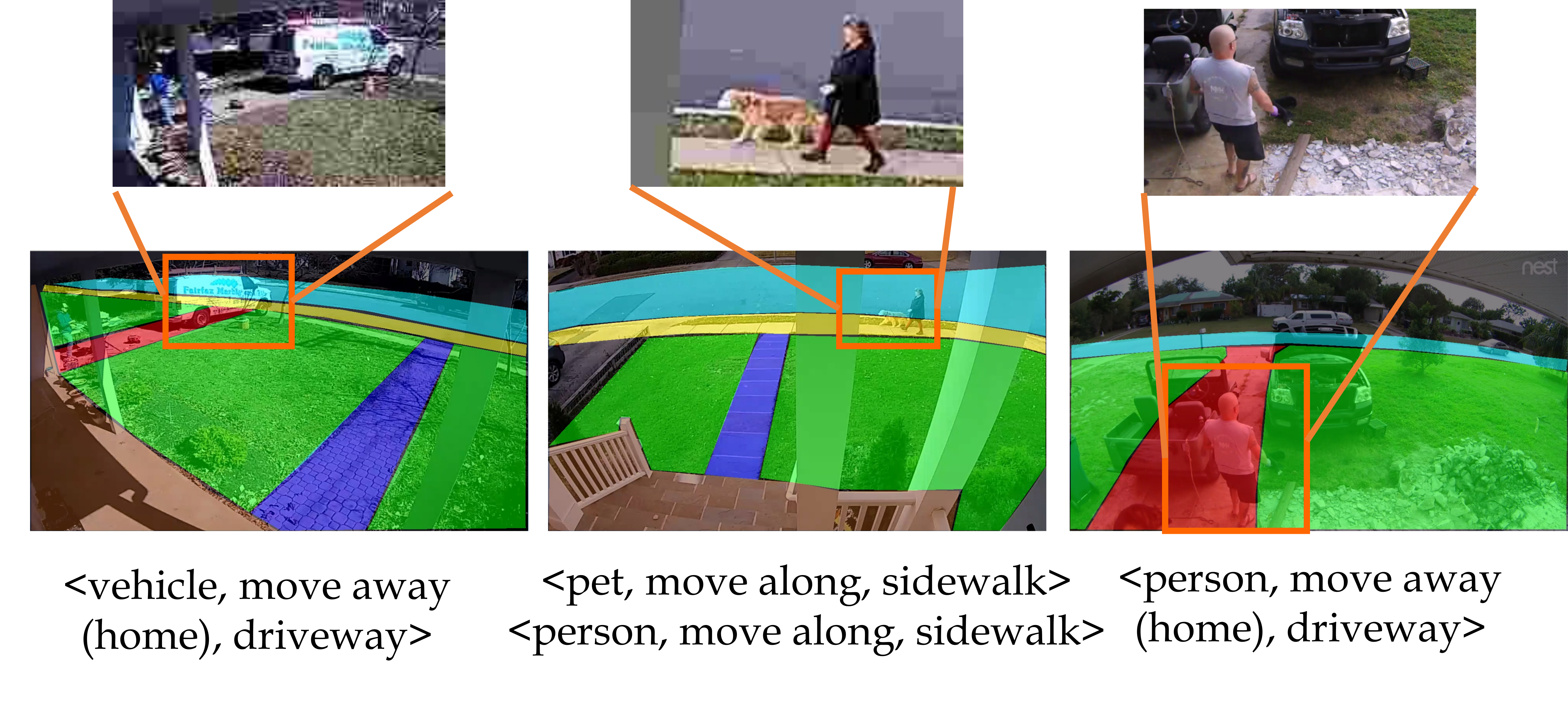}
  \caption{Example agent-in-place actions and segmentation maps. Different colors represent different places. We zoom in to the agents performing the actions for clarity. An agent-in-place action is represented as \textit{<agent, action, place>}. Same colors indicate same place types (\eg green for lawn, blue for walkway, \etc).
}
\label{fig:exp}
\end{figure}

Recent advances in deep neural networks have brought significant improvements to many fundamental computer vision tasks, including video action recognition \cite{twostream1,twostream2,twostream3,actionseg1,videodataset1,videodataset2,videodataset3,videodataset4}. Current action recognition methods are able to detect, recognize or localize general actions and identify the agents (people, vehicles, \textit{etc}.) \cite{twostream1,twostream2,twostream3,videodataset1,videodataset2,videodataset3,videodataset4,lstm1,lstm2,lstm3,C3D,C3D2,C3D3}. 
However, in applications such as surveillance, relevant actions often involve locations and moving directions that relate to to scene layouts --for example, it might be of interest to detect (and issue an alert about) a person walking towards the front door of a house, but not to detect a person walking along the sidewalk. So, what makes an action "interesting" is how the it interacts with the geometry and topology of the scene. 

Examples of these actions in outdoor home surveillance scenarios and the semantic segmentation maps of scenes are shown in Fig.\ref{fig:exp}. We refer to these actions as "\textit{agent-in-place}" actions to distinguish them from the widely studied generic action categories. From those examples we observe that
although the types of place are limited (\eg street, walkway, lawn), the layout (\ie structure of places, including reference to location, size, appearance of places and their adjacent places) vary significantly from scene to scene.
Without large-scale training data (which is hard to collect considering privacy issues), a naive method that directly learns from raw pixels in training videos without layout modeling can easily overfit to scene-specific textures and absolute pixel coordinates, and exhibit poor generalization on layouts of new scenes. 

\begin{figure}[!h]
\centering
  \includegraphics[width=\linewidth]{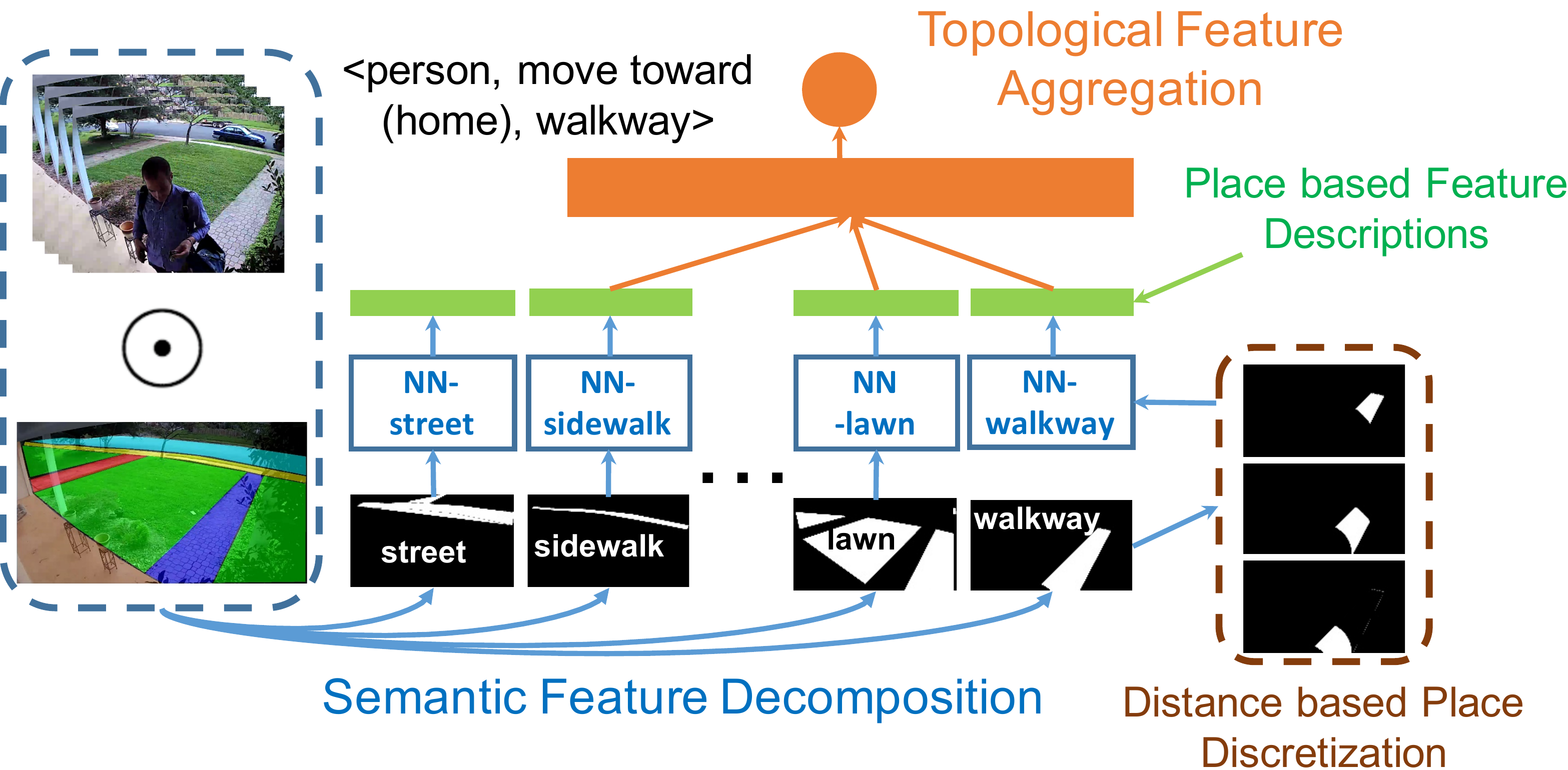}
  \caption{Framework of LIVR. Given the segmentation map, we decompose the semantic features into different places and extract place-based feature descriptions individually, then dynamically aggregate them at inference time according to the topology of the scene. $\odot$ denotes the masking operation for spatial decomposition. "NN" stands for neural network.
}
\label{fig:LIVR}
\end{figure}

To address the generalization problem, we propose the \textit{Layout-Induced Video Representation} (LIVR), which explicitly models scene layout for action recognition by encoding the layout in the network architecture given semantic segmentation maps.
The representation has three components:
\textit{1)} A semantic component represented by a set of bitmaps used for decomposing features in different \textit{"places"} (\eg walkway, street, \etc), and force the network to learn place-based features that are independent of scene layout;
\textit{2)} A geometric component represented by a set of coarsely quantized distance transforms of each semantic place incorporated into the network to model moving directions; 
\textit{3)} A topological component represented through the connection structure in a dynamically gated fully connected layer of the network--essentially aggregating representations from adjacent (more generally $h$-connected for $h$ hops in the adjacency graph of the semantic map) places. 
By encoding layout information (class membership of places, layout geometry and topology) into the network architecture using this decompotision-aggregation framework, we encourage our model to abstract away low-level appearance variations and focus on modeling high-level scene layouts, and eliminate the need to collect massive amounts of training data. 

\begin{figure}[!t]
\centering
\subfigure[Place Discretization]{\label{fig:DT_V}\includegraphics[width=0.45\linewidth]{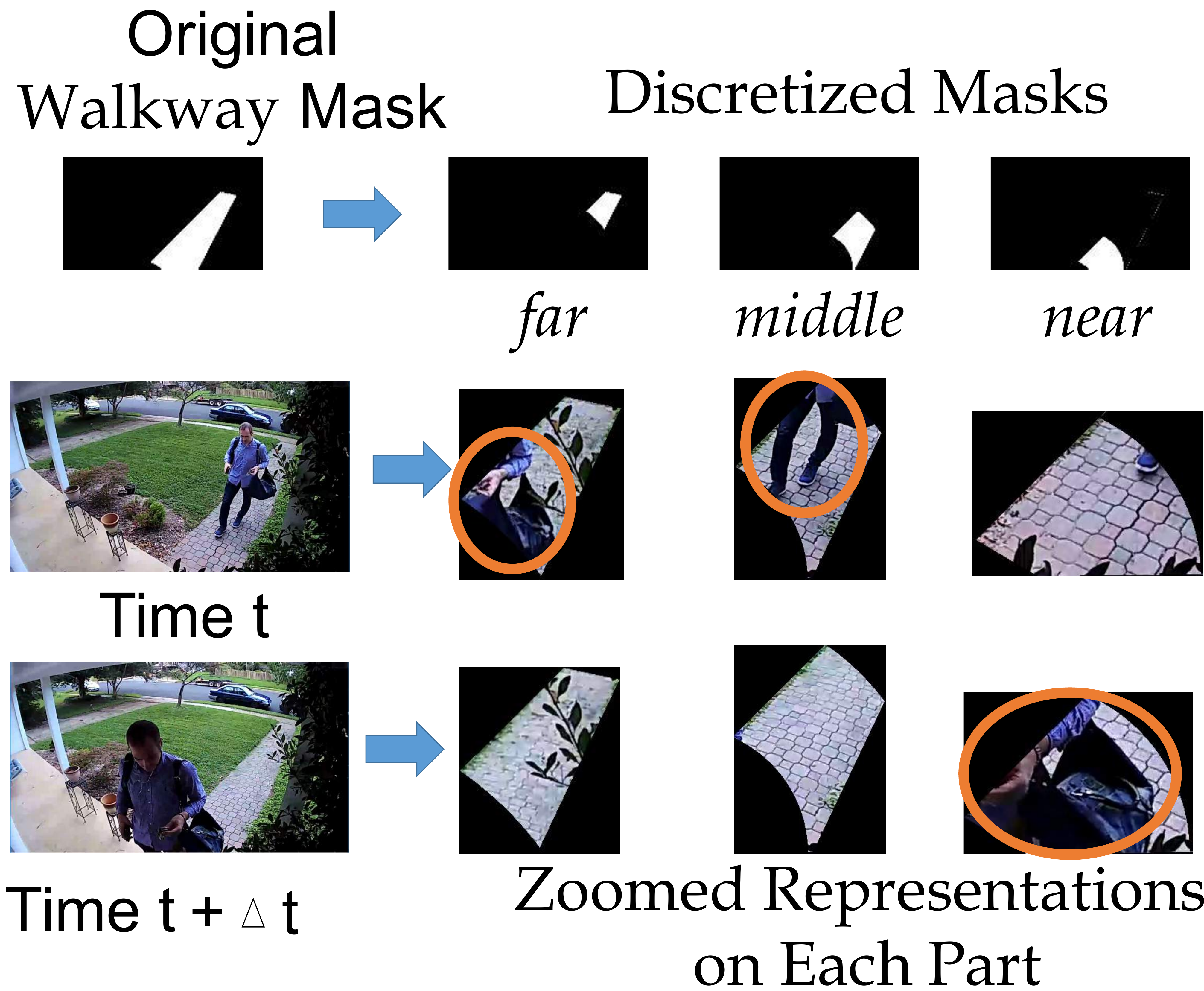}}\qquad
\subfigure[Topological Feature Aggregation]{\label{fig:topo_V}\includegraphics[width=0.45\linewidth]{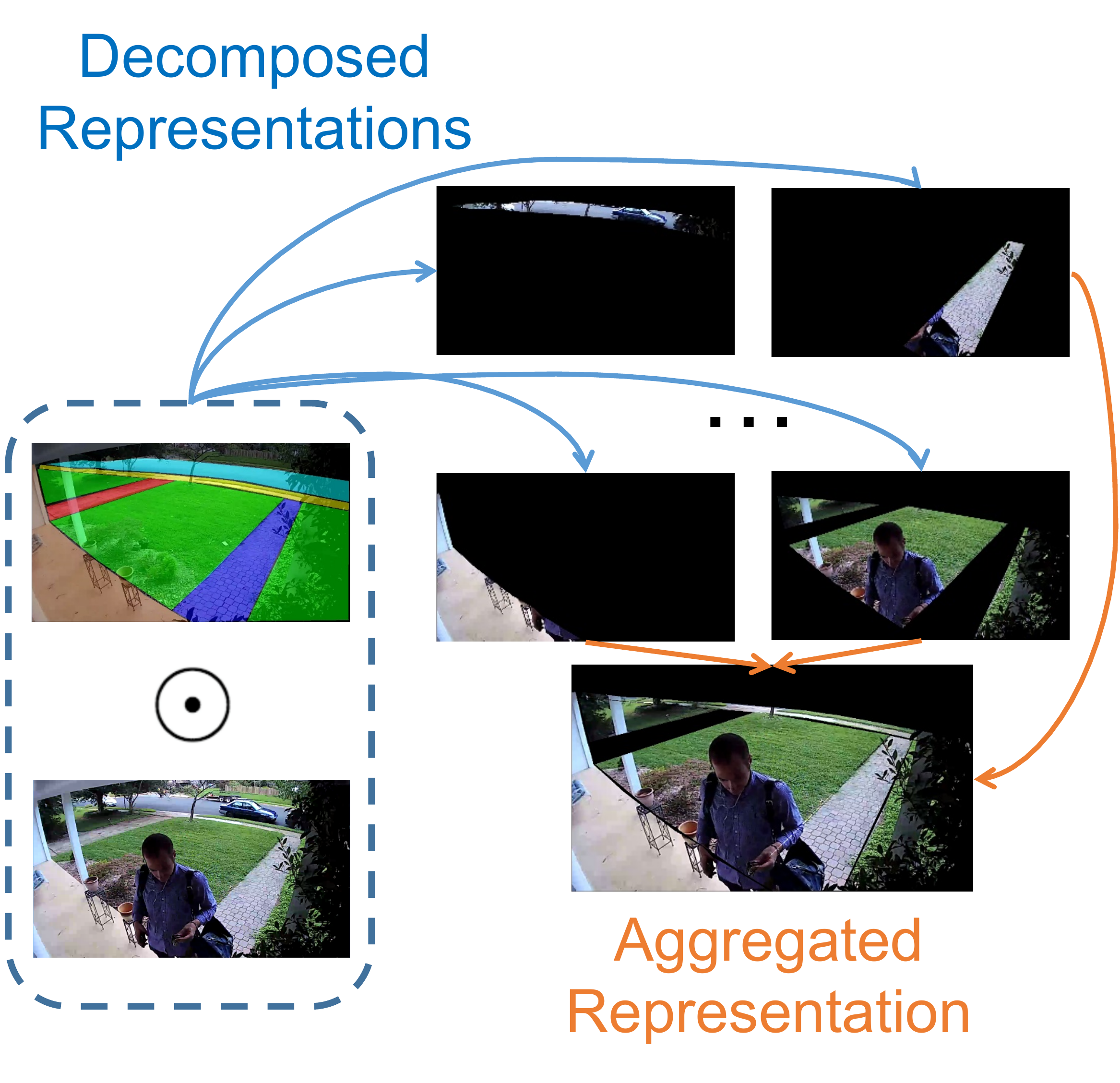}}
\caption{(a) illustrates distance-based place discretization. 
We segment the bit mask representing a given semantic class based on the distance transform of a second class, to explicitly represent the spatial-temporal order of moving agents which captures the moving direction w.r.t. that place. For example, this figure shows the partition of the \textit{walkway} map into components that are "far," "middle" and "near" to the porch class.
We use \textit{move toward (home)} action as an example: we first observe the person on the part of \textit{far} and \textit{middle} (distance), and after some time, the person appears in the part of \textit{near}. We use orange ellipses to highlight person parts. (b) illustrates the motivation behind topological feature aggregation.  We seek a representation that covers the entire body of the person, which can be accomplished by aggregating the masked images from places that are connected to \textit{walkway}.} 
\label{fig:illustration}
\end{figure}

The first two components require \textit{semantic feature decomposition} (Fig.\ref{fig:LIVR}). We utilize bitmaps encoded with the semantic labels of places to decompose video representations into different places and train models to learn place-based feature descriptions.
This decomposition encourages the network to learn features of generic place-based motion patterns that are independent of scene layouts. 
As part of the semantic feature decomposition, we encode scene geometry to model moving directions by discretizing a place into parts based on a quantized distance transform w.r.t. another place. Fig.\ref{fig:LIVR} (brown) shows the discretized bitmaps of \textit{walkway} w.r.t. \textit{porch}. As illustrated in Fig.\ref{fig:DT_V}, features decomposed by those discretized bitmaps capture moving agents in spatial-temporal order, which reveals the moving direction, and can be generalized to different scene layouts. 

The actions occurring in one place may be projected onto adjacent places from the camera view(see Fig.\ref{fig:topo_V}). We propose \textit{topological feature aggregation} to dynamically aggregate the decomposed features within the place associated with that action and adjacent places. The aggregation controls the "on/off" state of neuron connections from generic place-based feature descriptions to action nodes to model scene layout based on topological connectivity among places. 

We created the Agent-in-Place Action (APA) dataset, which to the best of our knowledge, is the first dataset that addresses recognizing actions associated with scene layouts. APA dataset contains over 5,000 15s videos obtained from 26 different surveillance scenes with around 7,100 actions from 15 categories. 
To evaluate the generalization of LIVR, we split the scenes into observed and unseen scenes. Extensive experiments show that LIVR significantly improves the generalizability of the model trained on only observed scenes and tested on unseen scenes (improving the mean average precision (mAP) from around 20\% to more than 50\%). Consistent improvements are observed on almost all action categories.

\section{Related Work}

\noindent\textbf{Video Action Recognition Methods and Datasets.}
Recent advances in video action recognition were driven by many large scale action recognition datasets. UCF101 \cite{videodataset2}, HMDB \cite{videodataset1} and Kinetics \cite{videodataset3} were widely used for recognizing actions in video clips \cite{C3D,p3d,NonLocal,twostream1,twostream2,twostream3,lstm1,lstm2,lstm3,C3D2,C3D3,closer}; THUMOS \cite{THUMOS}, ActivityNet\cite{videodataset4} and AVA \cite{AVA} were introduced for temporal/spatial-temporal action localization \cite{CDC,rc3d,attendandinteract,actorcentric,DTP,TemporalAD,sstad_buch_bmvc17,chao:cvpr2018,Lin:2017:SST:3123266.3123343}. Recently, significant attention has been drawn to model human-human \cite{AVA} and human-object interactions in daily actions \cite{mp2_cooking,charades,DAHLIA}. 
In contrast to these datasets that were designed to evaluate motion and appearance modeling, or human-object interactions,
our Agent-in-Place Action (APA) dataset is the first one that focuses on actions that are defined with respect to scene layouts, including interaction with places and moving directions. Recognizing these actions requires the network to not only detect and recognize agent categories and motion patterns, but also how they interact with the layout of the semantic classes in the scene. 
With the large variations of scene layouts, it is critical to explicitly model scene layout in the network to improve generalization on unseen scenes.

\vskip 0.5em\noindent\textbf{Surveillance Video Understanding.}
Prior work focuses on developing robust, efficient and accurate surveillance systems that can detect and track actions or events \cite{w4,surveillance1,surveillance12}, or detect abnormal events \cite{abnormal1,abnormal2,abnormal3}.
The most related work to ours is ReMotENet \cite{remotenet}, which skips expensive object detection \cite{faster,fast,yu4,yu1,angli} and utilizes 3D ConvNets to detect motion of an object-of-interest in outdoor home surveillance videos. We employ a similar 3D ConvNet model as proposed in \cite{remotenet} as a backbone architecture for extracting place-based feature descriptions for our model.



\vskip 0.5em\noindent\textbf{Knowledge Transfer.}
The biggest challenge of agent-in-place action recognition is to generalize a model trained with limited scenes to unseen scenes. Previous work on knowledge transfer for both images and videos has been based on visual similarity, which requires a large amount of training data \cite{KnowledgeTransfer,KT_image1,KT_image2,KT_image3,KT_video1,KT_video2,KT_video3}. For trajectory prediction, Ballan \etal \cite{KnowledgeTransfer} transferred the priors of statistics from training scenes to new scenes based on scene similarity. Kitani \etal \cite{activityforcasting} extracted static scene features to learn scene-specific motion dynamics for predicting human activities. 
Instead of utilizing low-level visual similarity for knowledge transfer, our video representation abstracts away appearance and location variance and models geometrical and topological relationships in a scene.
which are more abstract and easier to generalize from limited training scenes.

\section{Layout-Induced Video Representation}

\subsection{Framework Overview}
\begin{figure}[!t]
\centering
  \includegraphics[width=\linewidth]{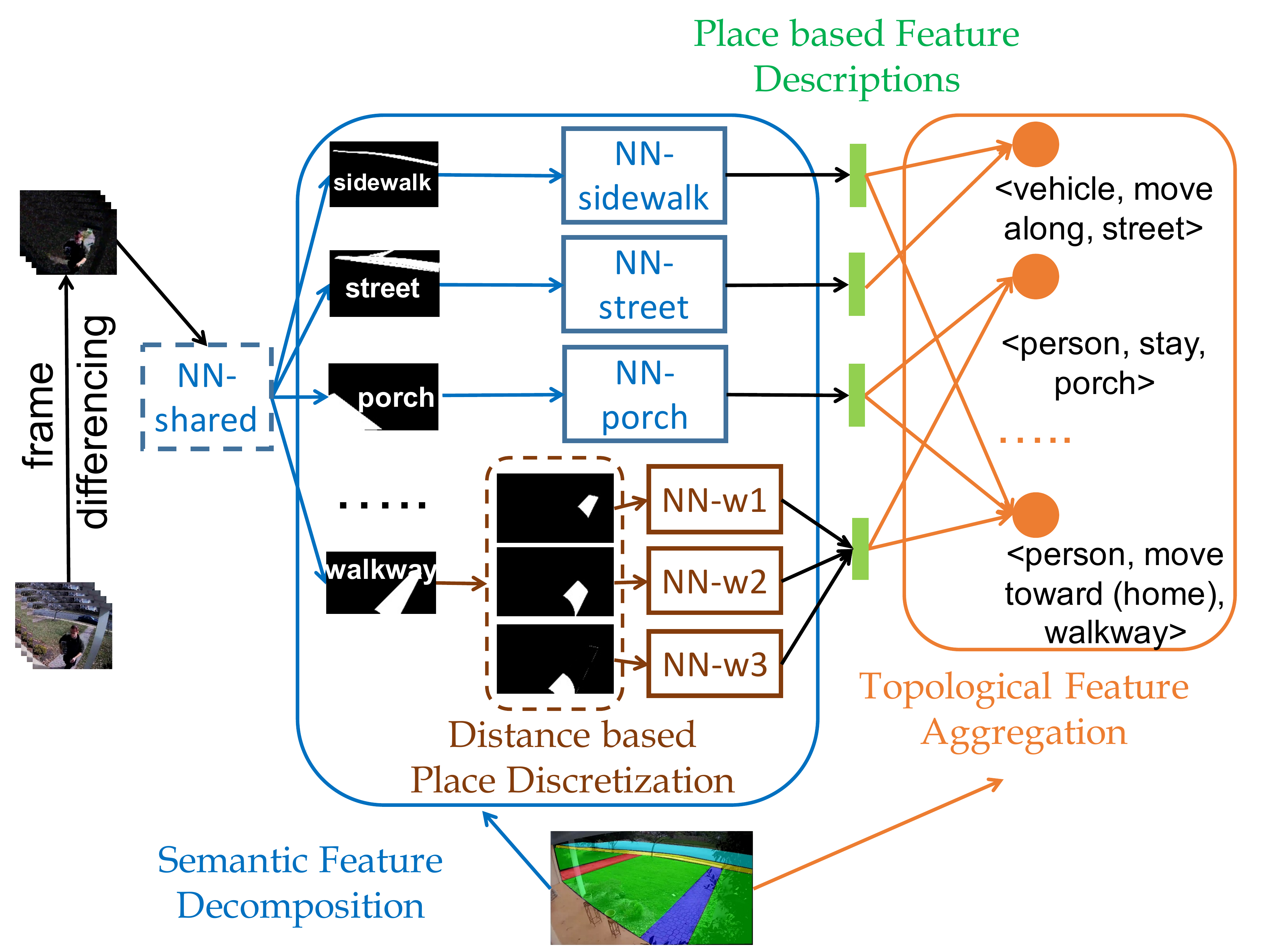}
  \caption{Layout-induced Video Representation Network: The dashed blue box indicates a shared 3D ConvNet to extract low-level features. We utilize the segmentation maps to decompose features into different places, and the solid blue boxes indicate that we train place-based models to extract place-based feature descriptions. When relevant to the activities of interest, we conduct distance-based place discretization to model moving directions; finally, we leverage the connectivity of places to aggregate the place-based feature descriptions at inference level.
}
\label{fig:sys}
\end{figure}

The network architecture of layout-induced video representation is shown in Fig.\ref{fig:sys}. For each video, we stack sampled frames of a video clip into a 4-D tensor. 
Our backbone network is similar to the architecture of ReMotENet \cite{remotenet}, which is composed of 3D Convolution (3D-conv) blocks. A key component of our framework is semantic feature decomposition, which decomposes feature maps according to region semantics obtained from given segmentation masks. This feature decomposition can be applied after any 3D-conv layer. 
Spatial Global Max Pooling (SGMP) is applied to extracted features within places, allowing the network to learn abstract features independent of shapes, sizes and absolute coordinates of both places and moving agents. For predicting each action label, we aggregate features from different places based on their connectivity in the segmentation map, referred to as Topological Feature Aggregation.

\subsection{Semantic Feature Decomposition}
\vskip 0.5em \noindent\textbf{Segmentation Maps}
Semantic Feature Decomposition utilizes a segmentation map of each place to decompose features and cause the network to extract place-based feature descriptions individually. 
The segmentation maps can be manually constructed using a mobile app we developed \footnote{Details for the app can be found in the supplementary materials.} to annotate each place by drawing points to construct polygons. This is reasonable since most smart home customers have only one or two cameras in their home.
We employ human annotations because automatic semantic segmentation methods segment places based on appearance (\eg color, texture, \etc), while our task requires differentiation between places with similar appearance based on functionality. 
For example, walkway, street and driveway have different functionalities in daily life, which can be easily and efficiently differentiated by humans. However, they may confuse appearance-based methods due to their similar appearance. Furthermore, since home surveillance cameras are typically fixed, users can annotate one map per camera very efficiently.
However, we will discuss the performance of our method using automatically generated segmentation maps in Sec. \ref{ablation}.


\vskip 0.5em \noindent\textbf{Place-based Feature Descriptions (PD).}
Given a segmentation map, we extract place-based feature descriptions as shown in the blue boxes in Fig.\ref{fig:sys}. 
We first use the segmentation map to decompose feature maps spatially into regions, each capturing the motion occurring in a certain place. The decomposition is applied to features instead of raw inputs to retain context information\footnote{An agent can be located at one place, but with part of its body projected onto another place in the view of the camera. If we use the binary map as a hard mask at input level, then for some places such as \emph{sidewalk}, \emph{driveway} and \emph{walkway}, only a small part of the moving agents will remain after the masking operation.}. Let $\mb{X}_L\in \mathbb{R}^{w_L\times\ h_L \times t_L \times c}$ be the output tensor of the $L^{th}$ conv block, where $w_L,h_L,t_L$ denote its width, height and temporal dimensions, and $c$ is the number of feature maps. The place-based feature description of a place indexed with $p$ is
\begin{equation}
\label{eq:func_de}
f_{L,p}(\mb X_L) = \mb{X}_L \odot \mathbb{I}\left[\mb{M}_L = p\right]
\end{equation}
where $\mb{M}_L\in \mathbb{I}^{w_L\times\ h_L \times 1}$ is the segmentation index map and $\odot$ is a tiled element-wise multiplication which tiles the tensors to match their dimensions. 
Place descriptions can be extracted from different levels of feature maps. $L = 0$ means the input level; $L > 0$ means after the $L^{th}$ 3D-conv blocks. A higher $L$ generally allows the 3D ConvNet to observe more context and to abstract features. We treat $L$ as a hyper-parameter and study its effect in Sec. \ref{EXPsec}.

\begin{figure}[h]
\centering
  \includegraphics[width=0.8\linewidth]{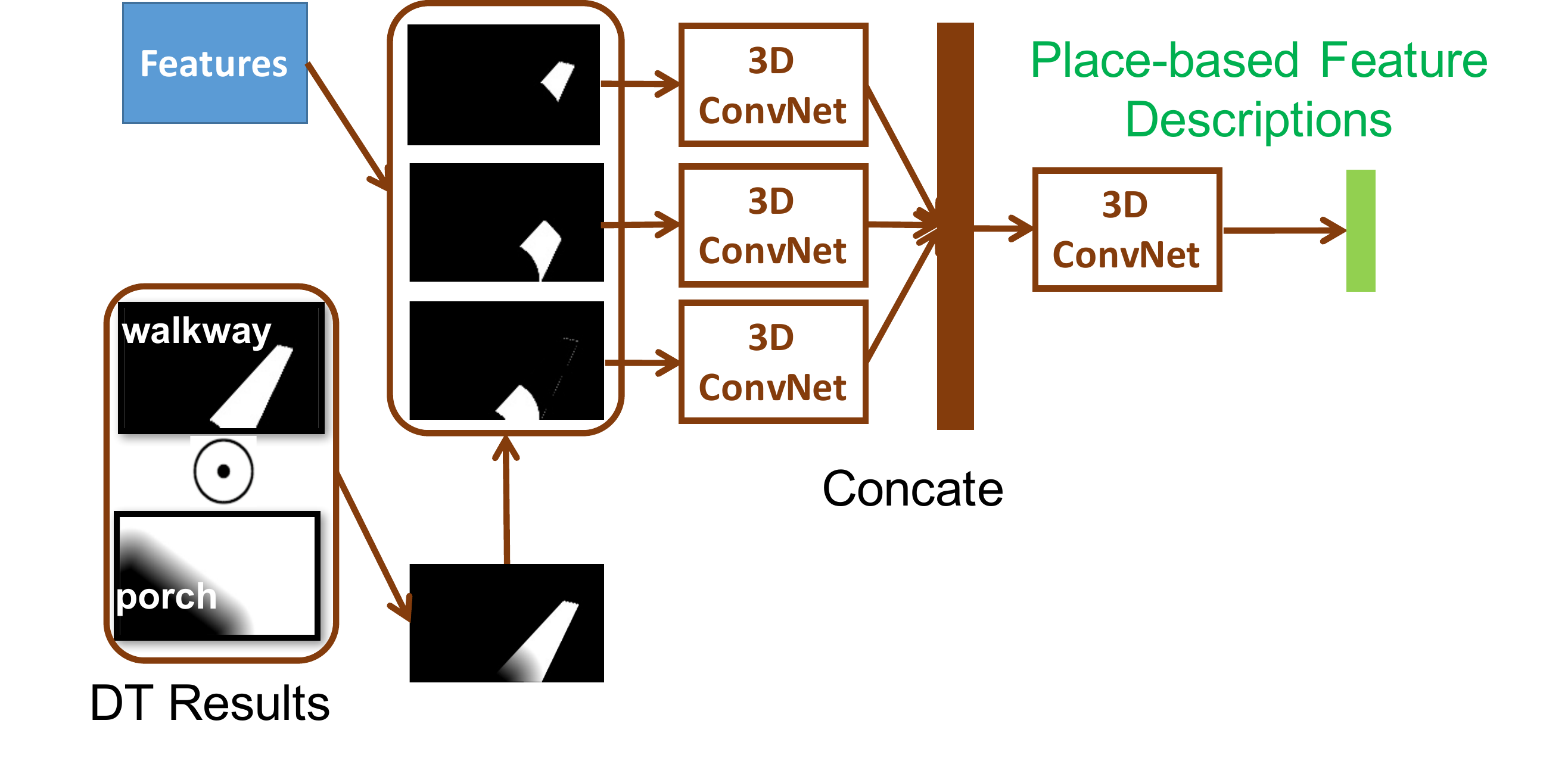}
  \caption{The process of distance-based place discretization.}
\label{fig:DT}
\end{figure}

\vskip 0.5em \noindent\textbf{Distance-based Place Discretization (DD).}
Many actions are naturally associated with moving directions \wrt some scene element (\eg the house in home surveillance). To learn general patterns of the motion direction in different scenes, we further discretize the place segmentation into several parts, and extract features from each part and aggregate them to construct the place-based feature description of this place. For illustration, we use \textit{porch} as the anchor place (shown in Fig.\ref{fig:DT}). We compute the distance between each pixel and the \emph{porch} in a scene (distance transform), and segment a place into $k$ parts based on their distances to \emph{porch}. 
The left bottom map in \ref{fig:DT} shows the porch distance transform of a scene. 
Let $D_L(x)$ be the distance transform of a pixel location $x$ in the $L^{th}$ layer. The value of a pixel $x$ in the part indexing map $\mb M^\Delta_L$ is computed as
\begin{equation}
M^\Delta_L(x)=\left\lfloor\frac{D_L^{\max}(x)-D_L^{\min}(x)}{k(D_{L}(x)-D_L^{\min}(x))}\right\rfloor
\end{equation}
where $D_L^{\max}(x)=\max\{D_L(x') | M_L(x')=M_L(x)\}$ and $D_L^{\min}(x)=\min\{D_L(x') | M_L(x')=M_L(x)\}$ are the max and min of pixel distances in the same place. They can be efficiently pre-computed.
The feature description corresponding to the $i^{th}$ part of $p^{th}$ place in $L^{th}$ layer is
\begin{equation}
    f^\Delta_{L,p,i}(\mb X_L)=\mb X_L\odot\mathbb{I}[\mb M_L=p\wedge \mb M^{\Delta}_L=i]
\end{equation}
where $\odot$ is the tiled element-wise multiplication.

Discretizing a place into parts at different distances to the anchor place and explicitly separating their spatial-temporal features allows the representation to capture moving agents in spatial-temporal order and extract direction-related abstract features.
However, not all places need to be segmented  since some places (such as \emph{sidewalk}, \emph{street}) are not associated with any direction-related action (\eg moving toward or away from the house). For these places, we still extract the whole-place feature descriptors $f_{L,p}$. We discuss the effect of different choices of place discretization and the number of parts $k$, and show the robustness of our framework to these parameters in Sec.~\ref{EXPsec}. 
To preserve temporal ordering, we apply 3D-conv blocks with spatial-only max pooling to extract features from each discretized place, and concatenate them channel-wise. Then, we apply 3D-conv blocks with temporal-only max pooling to abstract temporal information. Finally, we obtain a 1-D place-based feature description after applying GMP (see Fig.\ref{fig:DT}).  
The final description obtained after distance-based place discretization has the same dimensionality as non-discretized place descriptions.

\subsection{Topological Feature Aggregation (Topo-Agg)}

\begin{figure}[!t]
\centering
  \includegraphics[width=0.8\linewidth]{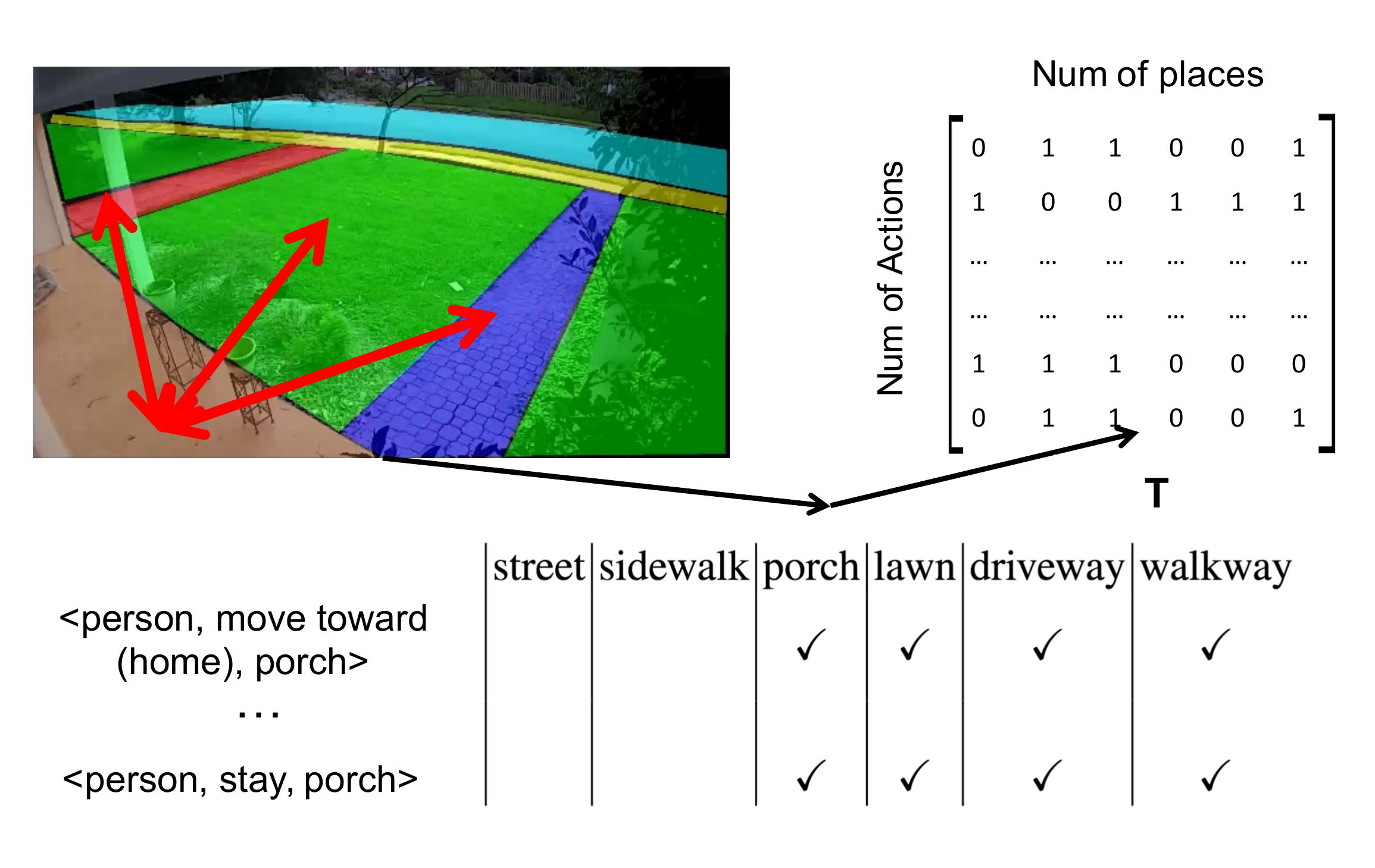}
  \caption{Topological feature aggregation which utilizes the connectivities between different places in a scene to guide the connections between the extracted place-based feature descriptions and the prediction labels. For clear visualization, we use the source places as \emph{ porch} in (b) with $h = 1$. The $\checkmark$ indicates we aggregate features from a certain place to infer the probability of an action.}
\label{fig:IM}
\end{figure}

Semantic feature decomposition allows us to extract a feature description for each place individually. To explicitly model the layout of a scene, we need to aggregate these place features
based on connectivity between places.
Each action category is one-to-one mapped to a place. To predict the confidence of an action $a$ occurring in a place $p$, features from adjacent places might provide contextual information, while the ones extracted far from place $p$ are distractors. To explicitly model the topological structure of places in a scene, we propose \textit{Topological Feature Aggregation}, which utilizes the spatial connectivity between places, to guide feature aggregation.

Specifically, as shown in Fig.\ref{fig:IM}, given a scene segmentation map, a source place $p$ and a constant $h$, we employ a Connected Component algorithm to find the $h$-connected set $C_h(p)$ which contains all places connected to place $p$ within $h$ hops.
The constant $h$ specifies the minimum number of steps to walk from the source to a destination place. 
Given the $h$-connected place set $C_h$, we construct a binary action-place matrix ($\mb{T}\in\mathbb{R}^{n_a\times n_p}$) for the scene where $n_a$ is the number of actions and $n_p$ is the number of places. $\mb{T}_{i,j}=1$ if and only if place $j$ is in the $C_h$ of the place corresponding to action $i$.
Fig.\ref{fig:IM} shows an example segmentation map with its action-place mapping, where $C_0(porch) = \{porch\}$, $C_1(porch) = \{porch, walkway, driveway, lawn\}$, $C_2(porch)$ includes all except for \emph{street}, and $C_3(porch)$ covers all six places. It is worth noting that since the vocabulary of our actions is closed, $\mb{T}$ is known at both training and testing time given the segmentation maps.


We implement topological feature aggregation using a gated fully connected layer with customized connections determined by the action-place mapping $\mb{T}$. Given $n_p$ $m$-D features extracted from $n_p$ places, we concatenate them to form a $(n_p\times m)$-D feature vector. We use $\mb{T}$ to determine the "on/off'' status of each connection of a layer between the input features and the output action nodes.
Let $\mb T_*=\mb T\otimes \mathbb{I}^{1\times m}$ be the actual mask applied to the weight matrix $\mb{W}\in \mathbb{R}^{n_a\times n_pm}$ where $\otimes$ is the matrix Kronecker product. The final output is computed by $\mb{y} = (\mb{W}\odot\mb{T}_*)\mb{f}_*$, 
where $\odot$ is element-wise matrix multiplication, $\mb{f}_*$ is the concatenated feature vector as the input of the layer. We omit bias for simplicity. 
Let $J$ be the training loss function (cross-entropy loss), considering the derivative of $\mb{W}$, the gradient formulation is $\nabla_{\mb{W}}J = (\nabla_{\mb{y}}J\mb{f}_*^\intercal)\odot \mb T_*$,
which is exactly the usual gradient ($\nabla_{\mb{s}}J\mb{f}^T$) masked by $\mb T_*$.
At training time, we only back-propagate the gradients to connected neurons. 

\section{Agent-in-Place Action Dataset}
We introduce a video dataset for recognizing agent-in-place actions.
We collected outdoor home surveillance videos from internal donors and webcams
to obtain over $7,100$ actions from around $5,000$ 15-second video clips with $1280\times 720$ resolution. These videos are captured from 26 different outdoor cameras which cover various layouts of typical American front and back yards. 

We select 15 common agent-in-place actions to label and each is represented as a tuple containing an action, the agent performing it, and the place where it occurs. The agents, actions, and places involved in our dataset are:
$agent=\{$\emph{person}, \emph{vehicle}, \emph{pet}$\}$;  $action=\{$\emph{move along}, \emph{stay}, \emph{move away (home)}, \emph{move toward (home)}, \emph{interact with vehicle}, \emph{move across}$\}$;  $place=\{$\emph{street}, \emph{sidewalk}, \emph{lawn}, \emph{porch}, \emph{walkway}, \emph{driveway}$\}$.

The duration of each video clip is 15s, so multiple actions can be observed involving one or more agents in one video. We formulate action recognition as a multi-label classification task. 
We split the 26 cameras into two sets: observed scenes (5) and unseen scenes (21) to balance the number of instances of each action in observed and unseen scenes and at the same time cover more scenes in the unseen set. We train and validate our model on observed scenes, and test its generalization capability on the unseen scenes. 
Details about the APA dataset including statistics can be found in the supplementary material.
\section{Experiments}\label{EXPsec}

\begin{table*}[!t]
\centering
\tiny
\caption{\textbf{The Path from Traditional 3D ConvNets to our Methods.} B/L1 and B/L2 are baseline models with raw pixels and and ConcateMap as input, respectively. For our proposed models: V1 uses segmentation maps to extract place-based feature descriptions only. V3 applies distance-based place discretization for some places. Both V1 and V3 use a FC layer to aggregate place features; V2 and V4 uses topological feature aggregation. H and FPS2 indicates using higher resolutions and FPS, and MF means using more filters per conv layer. Besides our baselines, we also compare LIVR with two state-of-the-art action recognition methods: \cite{remotenet,tsn}.}
\label{table:performance}
\scriptsize
\begin{tabular}{@{}c|ccc|cccc|ccc|cc@{}}
\toprule
Network Architecture & \begin{tabular}[c]{@{}c@{}}B/L1\end{tabular} & \begin{tabular}[c]{@{}c@{}}B/L2\end{tabular} & \begin{tabular}[c]{@{}c@{}}B/L2 \\+MF\end{tabular}& \begin{tabular}[c]{@{}c@{}}LIVR-\\V1\end{tabular}        & \begin{tabular}[c]{@{}c@{}}LIVR-\\V2\end{tabular} &  \begin{tabular}[c]{@{}c@{}}LIVR-\\V3\end{tabular} & 
\begin{tabular}[c]{@{}c@{}}LIVR-\\V4\end{tabular} & \begin{tabular}[c]{@{}c@{}}LIVR-\\V4+H\end{tabular} & \begin{tabular}[c]{@{}c@{}}LIVR-\\V4+MF\end{tabular} &
\begin{tabular}[c]{@{}c@{}}LIVR-\\V4+FPS2\end{tabular} &
\begin{tabular}[c]{@{}c@{}}TSN \\\cite{tsn}\end{tabular} &
\begin{tabular}[c]{@{}c@{}}ReMotENet \\\cite{remotenet}\end{tabular}\\

\toprule
3D ConvNet?         & \checkmark                                                                                        & \checkmark                                                                                                       & \checkmark                                         & \checkmark                                             & \checkmark                                             & \checkmark                                                 & \checkmark                                                                 & \checkmark               &\checkmark                                                    &\checkmark  &-&-\\
ConcateMap?          &                                                                                                                  & \checkmark          &\checkmark                                                                                                       &                                                    &                                                        &                                                        &                                                            &                                                                            &                                                                             &&- &- \\
place-based feature description?                                                               &                                                              &           &                                                                                                       & \checkmark                                         & \checkmark                                             & \checkmark                                             & \checkmark                                                 & \checkmark                                                                 & \checkmark                                                                  &\checkmark   &- &- \\
distance-based place discretization?                                                               &                                                              &        &                                                                    &                                &                                             & \checkmark                                                 & \checkmark                                                                 & \checkmark                                                                & \checkmark  &\checkmark    &- &- \\
topological feature aggregation?                                                                  &                                                              &      &                                                                      &            & \checkmark                                                                                  &                                                        & \checkmark                                                 & \checkmark                                                                 & \checkmark                                                                 &\checkmark    &- &- \\
higher resolutions?                                        &                       &                                                              &                                                                                                                      &                                                    &                                                        &                                                        &                                                            & \checkmark                                                                 &                                                                             & & -&- \\
more filters?                                                               &                                                              &             & \checkmark                                                                                                                   &                                                    &                                                        &                                                        &                                                            &                                                                            & \checkmark                                                                  &  &- &- \\
higher FPS?                                   &                           &                                                              &                                                                                                                                &                                                    &                                                        &                                                        &                                                            &                                                                            &                                                                & \checkmark   &- &- \\
\midrule
Observed scenes mAP      & 51.09                                                               & 54.12         &        53.02                                      & 55.69      & 57.12                                                                                        & 58.02                                                & \textbf{59.71}                                                  & 59.64                                                        & 59.52                                                 & 59.01  &56.71 & 55.92\\
Unseen scenes mAP & 19.21                                                               & 21.16                &     20.45                                    & 41.57      & 43.78                                                                                          & 47.76                                                & 50.65                                                    & 49.03                                                         & \textbf{50.98}           &      49.56                                       
 &23.21 &22.05 \\ \bottomrule                                                   
\end{tabular}
\end{table*}

\subsection{Implementation Details}
\noindent
\textbf{Network Architecture.}
Unlike traditional 3D ConvNets which conduct spatial-temporal max-pooling simultaneously, we found that decoupling the pooling into spatial-only and temporal-only leads to better performance (experimental results and discussions can be found in the supplementary materials). So, for each place-specific network that extracts place-based feature descriptions, we utilize nine blocks of 3D ConvNets with the first five blocks using spatial-only max pooling and the last four blocks using temporal-only max pooling.  The first two blocks have one 3D-conv layer each, and there are two convolutional (conv) layers with ReLU in between for the remaining blocks. 
For each place-specific network, we use 64 $3\times3\times3$ conv filters per 3D-conv layer. After conducting SGMP on features extracted by each place-specific network, the final concatenated 1-D feature dimension is $6\times64$ since there are 6 places in total. The inference is conducted with a gated fully connected layer, whose connections ("on/off" status) are determined by action labels and scene topology. We use the sigmoid function to obtain the predicted probability of each action. We decompose semantics after the second conv blocks ($L=2$); we conduct distance-based place discretization on $PL_{DT} = \{\emph{\text{walkway}}, \emph{\text{driveway}}, \emph{\text{lawn}}\}$ and choose $k=3$; for topological feature aggregation, we choose $h=1$. 
The detailed place-specific network structure is shown in the supplementary materials.

\noindent
\textbf{Anchor Place.}
For our dataset, the directions mentioned are all relative to the house location, and \emph{porch} is a strong indicator of the house location. So we only conduct distance transform to \emph{porch}
\footnote{If there is no \emph{porch} in a scene, we let the user to draw a line (click to generate two endpoints) to indicate its location.}\nopagebreak, 
but the distance-based place discretization method can be easily generalized to represent moving direction w.r.t any arbitrary anchor place.

\begin{figure*}[!h]
\centering
  \includegraphics[width=0.9\linewidth]{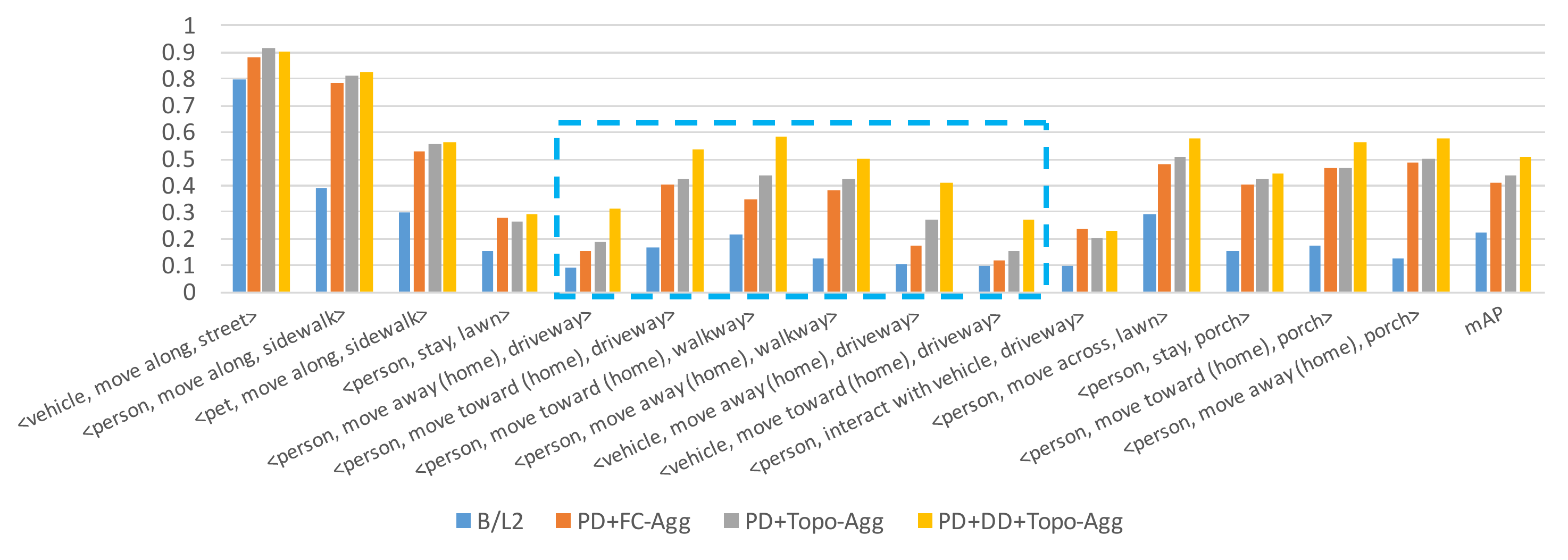}
  \caption{Per-category average precision of the baseline 3 and our methods on unseen scenes.
The blue dashed box highlights actions which require modeling moving directions. We observe that the proposed place-based feature descriptions (PD), distance-based place discretization (DD) and topological feature aggregation (Topo-Agg) significantly improve the average precision on almost all action categories. FC-Agg stands for using a FC layer to aggregate place descriptions.}
\label{fig:perCate}
\end{figure*}

\begin{figure}[!h]
\centering
  \includegraphics[width=\linewidth]{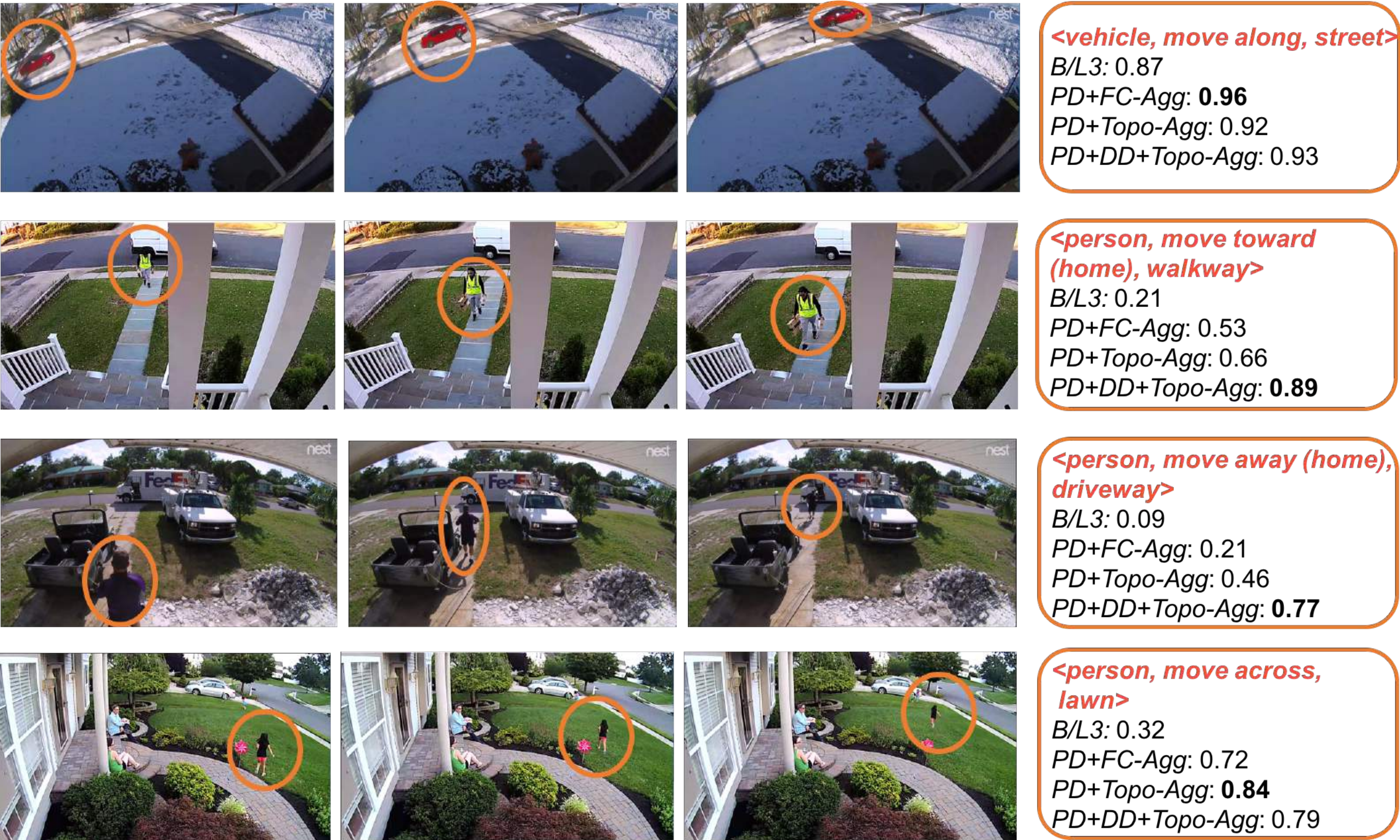}
  \caption{Qualitative examples: The predicted confidences of groundtruth actions using different methods. We use 3 frames to visualize a motion and orange ellipses to highlight moving agents.}
\label{fig:V}
\end{figure}

\noindent
\textbf{Training and Testing Details.}
Our action recognition task is formulated as multi-label classification without mutual exclusion. 
The network is trained using the Adam optimizer \cite{adam} with 0.001 initial learning rate.
For input video frames, we follow \cite{remotenet} to use FPS 1 and down-sample each frame to $160\times90$ to construct a $15\times160\times90\times3$ tensor for each video as input. Suggested by \cite{remotenet}, small FPS and low resolution are sufficient to model actions for home surveillance where most agents are large and the motion patterns of actions are relatively simple. 
We evaluate the performance of recognizing each action independently and report Average Precision (AP) for each action and mean Average Precision (mAP) over all categories. 

\noindent
\textbf{Dataset Split.}
We split the 26 scenes into two sets: observed scenes and unseen scenes. We further split the observed scenes into training and validation sets with a sample ratio of nearly $1:1$. The model is trained on observed scenes and test on unseen scenes. The validation set is used for tuning hyperparameters, which are robust with different choices (see Sec. \ref{ablation}).

\subsection{Baseline Models}
We follow \cite{remotenet} to employ 3D ConvNets as our baseline (B/L) model. 
The baseline models share the same 3D ConvNets architecture with our proposed model, except that the last layer is fully connected instead of gated through topological feature aggregation. The difference between baselines is their input: B/L1 takes the raw frames as input; B/L2 incorporates the scene layout information by directly concatenating the 6 segmentation maps to the RGB channels in each frame (we call this method ConcateMap), resulting in an input of 9 channels per frame in total.
We train the baseline models using the same setting as in the proposed model, and the performance of the baselines are shown in column 2-5 in Table \ref{table:performance}. We observe that: 1) the testing performance gap between observed and unseen scenes is large, which reveals the poor generalization of the baseline models; 2) marginal improvements are obtained by incorporating scene layout information using ConcateMap, which suggests that it is difficulty for the network to learn the human-scene interactions directly from the raw pixels and segmentation maps.
In addition, we also train a B/L2 model with $6\times$ more filters per layer to evaluate whether model size is the key factor for the performance improvement. The result of this enlarged B/L2 model is shown in column 5 of Table \ref{table:performance}.
Overall, the baseline models which directly extract features jointly from the entire video suffer from overfitting, and simply enlarging the model size or directly using the segmentation maps as features does not improve their generalization.
More details about baseline models can be found in the supplementary materials.

\subsection{Evaluation on the Proposed Method}
\noindent\textbf{The path from traditional 3D ConvNets to our method.}
We show the path from the baselines to our method in Table \ref{table:performance}. 
In column 6-9, we report the mAP of our models on observed scene validation set and unseen scene testing set. We observe three significant performance gaps, especially on unseen scenes: 1) from B/L2 to LIVR-V1, we obtain around $20\%$ mAP improvement by applying the proposed semantic feature decomposition to extract place feature descriptions; 2) from LIVR-V1 to LIVR-V3, our model is further improved by explicitly modeling moving directions by place discretization; 3) when compared to using a fully connected layer for feature aggregation (V1 and V3), our topological method (V2 and V4) leads to another significant improvement, which shows the efficacy of feature aggregation based on scene layout connectivity. 
We also evaluate the effect of resolutions, FPS and number of filters using our best model (LIVR-V4). Doubling the resolution ($320\times180$), FPS (2) and number of filters (128) only results in a slight change of the model's accuracy (columns 10-12 in Table \ref{table:performance}). 
Besides our baselines, we also apply other state-of-the-art video action recognition methods (TSN \cite{tsn} and ReMotENet \cite{remotenet}) on our dataset. LIVR outperforms them by a large margin, especially on the unseen scenes.

\noindent\textbf{Per-category Performance.}
Fig.\ref{fig:perCate} shows the average precision for each action on unseen scenes. LIVR outperforms the baseline methods by a large margin on almost all action categories.
When comparing the orange and green bars in Fig.\ref{fig:perCate}, we observe that the proposed topological feature aggregation (Topo-Agg) leads to consistently better generalization for almost all actions. 
The blue dashed box highlights the actions that include moving directions, and consistent improvements are brought by distance-based place discretization (DD). 
For some actions, especially the ones occurring on street and sidewalk, since they are relatively easy to recognize, adding DD or Topo-Agg upon the place-based feature descriptions (PD) does not help much.
Overall, LIVR improves the generalization capability of the network, especially for actions that are more challenging, and are associated with moving directions.

\noindent\textbf{Qualitative Results.}
Some example actions are visualized using three frames in temporal order and the predicted probabilities of the groundtruth actions using different methods are reported in Fig.\ref{fig:V}. It is observed that for relatively easy actions such as \emph{<vehicle, move along, street>}, performance is similar across approaches. However, for challenging actions, especially ones requiring modeling moving directions such as \emph{<person, move toward (home), walkway>}, our method outperforms baselines significantly.

\begin{figure*}[!h]
\centering
  \subfigure[Decomposition Level $L$]{\label{fig:allLevel}\includegraphics[width=0.23\linewidth]{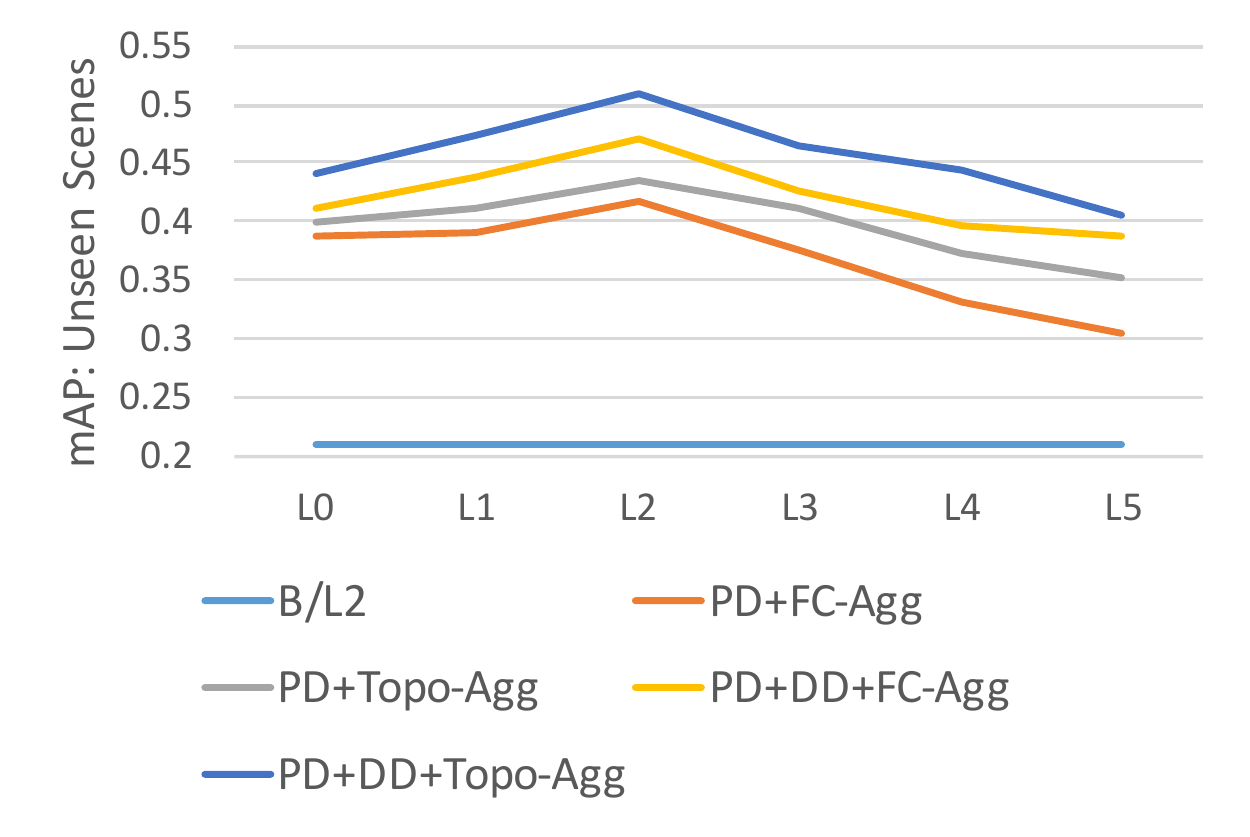}}
  \subfigure[Place Discretization]{\label{fig:DT-choice}\includegraphics[ width=0.23\linewidth]{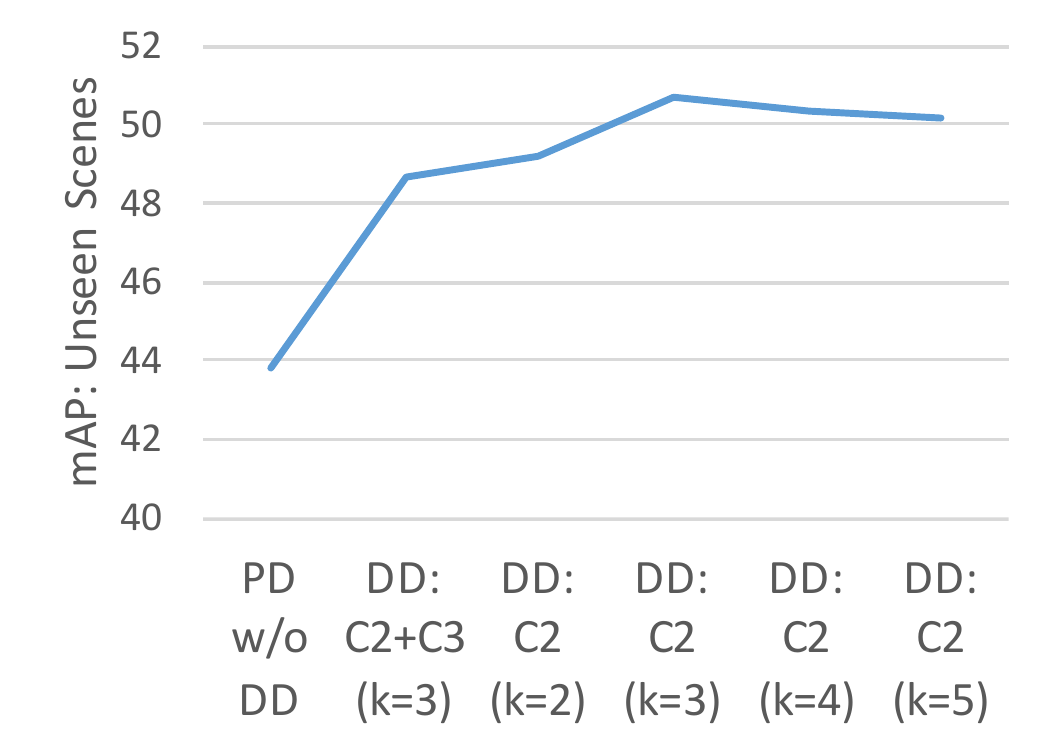}}
\subfigure[Topological Aggregation]{\label{fig:topo}\includegraphics[ width=0.23\linewidth]{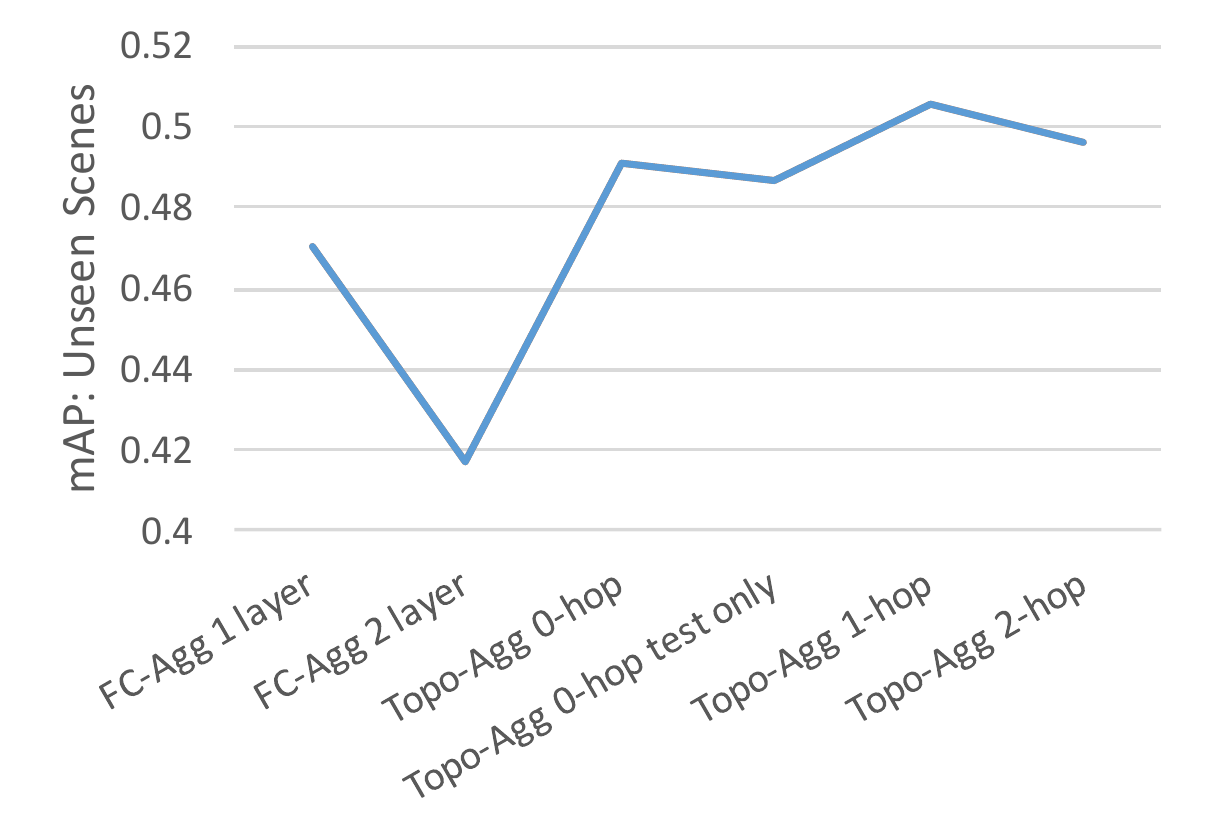}}
\subfigure[Segmentation: GT v.s. Auto]{\label{fig:auto_performance}\includegraphics[ width=0.23\linewidth]{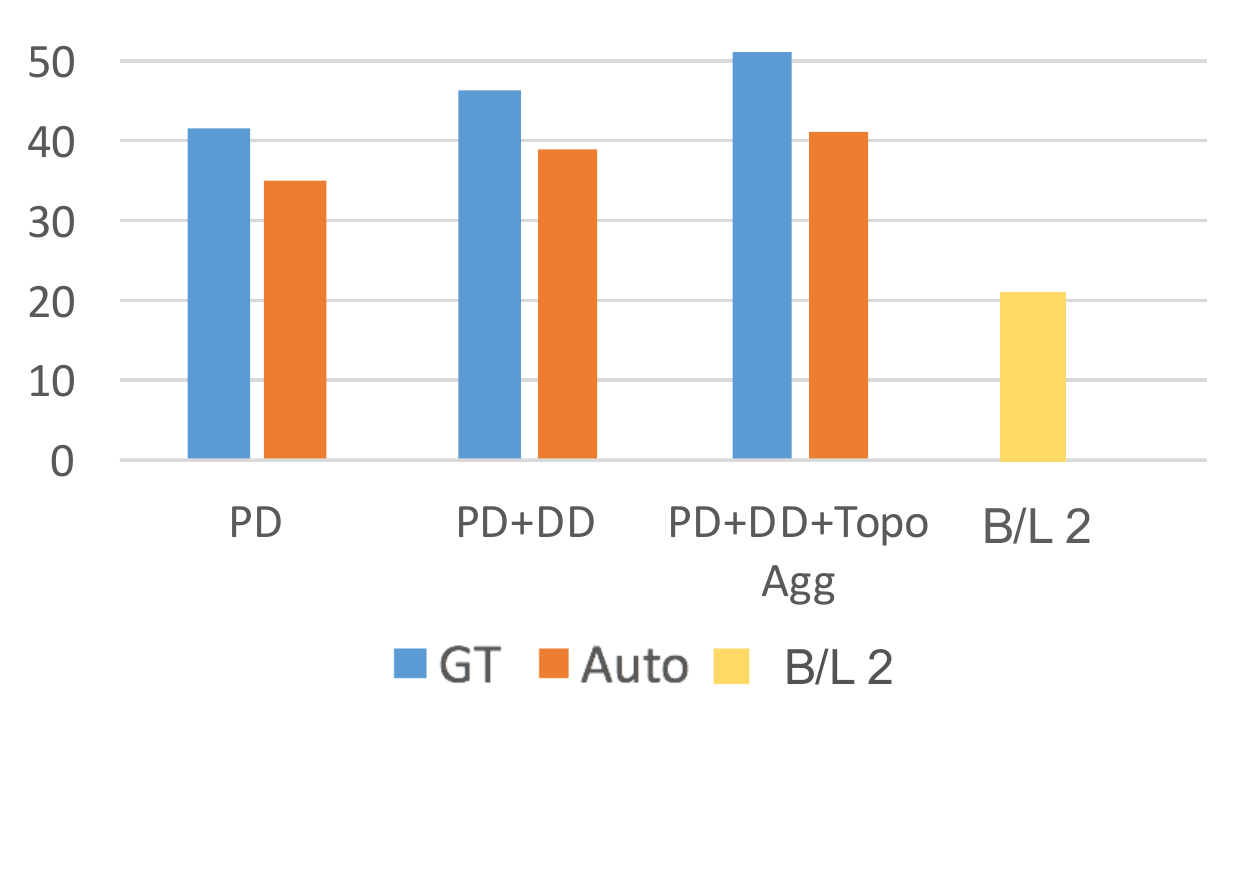}}
  \caption{Evaluation: (a) The effect of extracting place-based feature descriptions (PD) at different levels using different variants of our proposed model. (b) Different strategies for distance-based place discretization. (c) Different feature aggregation approaches on unseen scenes. (d) Performance of LIVR using groundtruth (GT) and automatically generated (Auto) segmentation map. }
\label{fig:eval1}
\end{figure*}

\subsection{Ablation Analysis on Unseen Scenes}\label{ablation}
\noindent\textbf{Place-based Feature Description.}
The hyper-parameter for PD is the level $L$, controlling when to decompose semantics in different places.
Fig.\ref{fig:allLevel} and \ref{fig:topo} show that the generalization capability of our model is improved when we allow the network to observe the entire video at input level, and decompose semantics at feature level (after the 2nd conv blocks). Generally, the improvements of PD are robust across different feature levels.

\begin{figure}[!t]
\centering
  \includegraphics[width=0.9\linewidth]{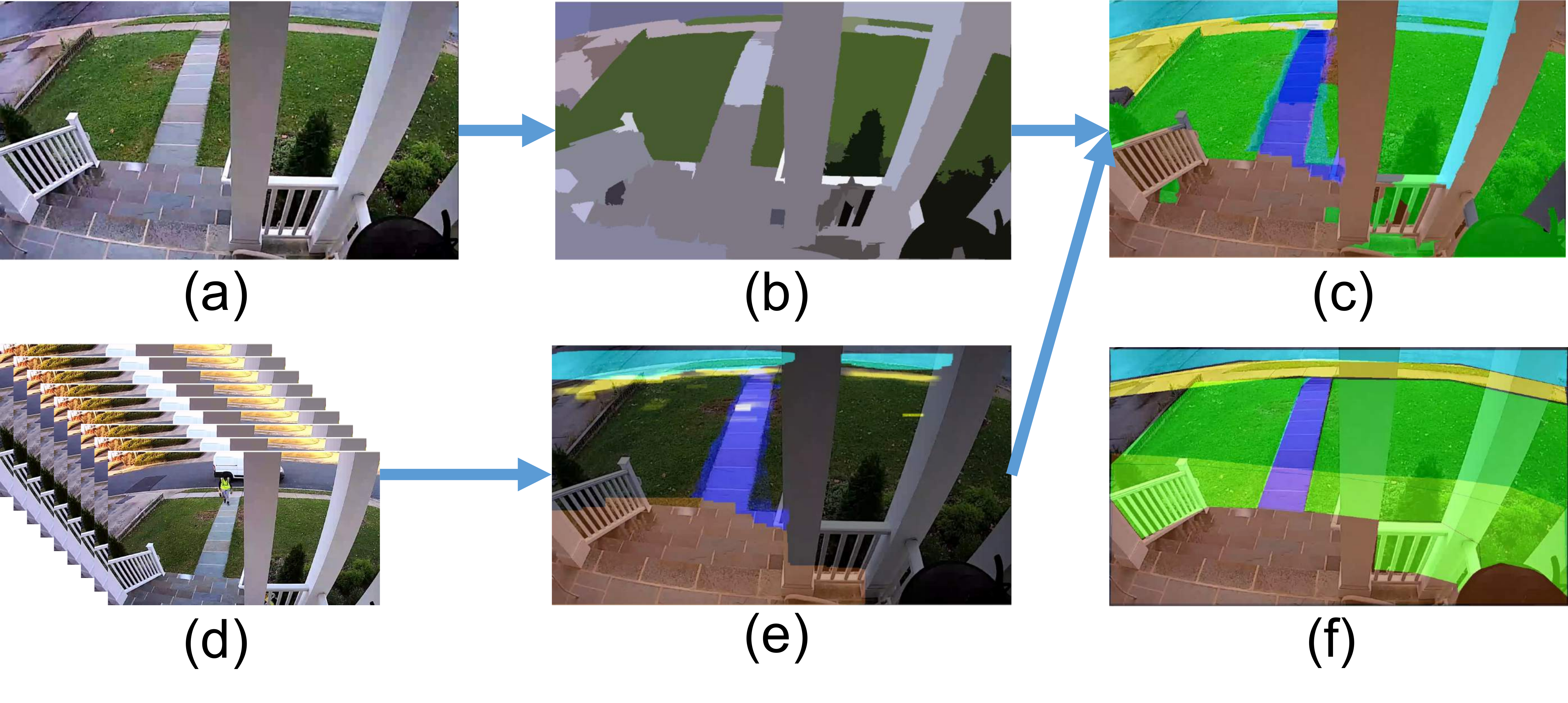}
  \caption{Process of Automatically Generating Segmentation Maps. (a) is the input camera image. (b) is the output of normalized cut method. (d) is the set of all videos captured by this camera. (e) shows the heatmaps we obtained by analyzing the patterns of moving objects from the videos. (c) is the generated segmentation map. (f) is the ground truth map.}
\label{fig:auto_main}
\end{figure}

\noindent\textbf{Distance-based Place Discretization.}
We study different strategies for determining $PL_{DT}$ and the number of parts to discretize ($k$) per place. From our observations, including the anchor place--\textit{porch}, the six places in our dataset can be clustered into three categories with regard to the distance to camera: C1 includes only \emph{porch}, which is usually the closest place to camera; C2 includes \emph{lawn}, \emph{walkway}, \emph{driveway}, and actions occurring in those places usually require modeling the moving direction directly; C3 includes \emph{sidewalk} and \emph{street}, which are usually far away from the house, and actions on them are not sensitive to directions (\eg "\textit{move along}"). 
We evaluate our method with two strategies to apply DD on: 1) all places belong to C2 and C3; 2) only places in C2. The results are shown in Fig.\ref{fig:DT-choice}. We observe that applying DD on C3 dose not help much, but if we only apply DD on places in C2, our method achieves the best performance. In terms of the number of discretized parts $k$, we evaluate $k$ from 2 to 5 and observe from Fig.\ref{fig:DT-choice} that the performance is robust when $k\geqslant3$.

\noindent\textbf{Topological Feature Aggregation.}
We evaluate different $h$ values to determine the $h$-connected set and different strategies to construct and utilize the action-place mapping $\mb{T}$. The results are shown in Fig.\ref{fig:topo}. We set $L=2$, and use both PD and DD. We observe that Topo-Agg achieves its best performance when $h=1$, \ie for an action occurring in place $P$, we aggregate features extracted from place $P$ and its directly connected places. In addition, we compare Topo-Agg to the naive fully connected inference layer (FC-Agg: 1 layer) and two fully-connected layers with 384 neurons each and a ReLU layer in between (FC-Agg: 2 layers).  Unsurprisingly, we observe that the generalizability drops significantly with an extra fully-connected layer, which reflects overfitting. Our Topo-Agg outperforms both methods.
We also conduct an experiment where we train a fully connected inference layer and only aggregate features based on topology at testing time (``Topo-Agg: $1$-hop test only'') and it shows worse performance. 

\noindent\textbf{LIVR with Automatically Generated Segmentation Maps.}
To evaluate the performance of LIVR using imperfect segmentation maps, we developed an algorithm to automatically generate the segmentation maps. As shown in Fig.\ref{fig:auto_main}, we first apply normalized cut \cite{Ncut} on the camera images to obtain segments (Fig.\ref{fig:auto_main} (b))\footnote{We also tried deep learning based methods trained on semantic segmentation datasets, but they perform poorly on our camera images. Details can be found in the supplementary materials.}. 
Then, to further differentiate different places with similar appearance (\eg walkway and street), we developed an algorithm to utilize the historical statistics obtained from previous videos (Fig.\ref{fig:auto_main} (d)) of a scene to generate heatmaps of some specific places\footnote{We utilize the patterns of moving objects to differentiate places. For example, street is a place where vehicles move with limited scale changes (from the camera perspective), and walkway is a place where people with notably large scale changes walk.}
(Fig.\ref{fig:auto_main} (e)). Then, the two results are combined to obtain final segmentation maps ((Fig.\ref{fig:auto_main} (c))). Our method can generate reasonably good segmentation maps when compared to the groundtruth maps obtained manually (Fig. \ref{fig:auto_main} (f)). We evaluate LIVR using the imperfect maps and observe some performance degradations (around $10\%$), but LIVR still outperforms the baselines by a large margin (around $20\%$), which demonstrate the effectiveness of our method even if the segmentation maps are imperfect. 
Details of how we generate the segmentation maps can be found in the supplementary materials.

\section{Conclusion}
To improve the generalization of a deep network that learns from limited training scenes, we explicitly model scene layout in a network by using layout-induced video representations, which abstracts away low-level appearance variance but encodes the semantics, geometry and topology of scene layouts.
An interesting future directions would be to include integrate the estimation of the semantic maps into the network architecture, which may require collecting more scenes for training. 

{\small
\bibliographystyle{ieee}
\bibliography{egbib}
}

\end{document}